\documentclass[runningheads]{llncs}
\usepackage{graphicx}

\usepackage{tikz}
\usepackage{comment}
\usepackage{amsmath,amssymb} 
\usepackage{color}
\usepackage{ifthen}

\usepackage[accsupp]{axessibility}  

\usepackage{dsfont}
\usepackage{multirow}
\usepackage{pgfplots}
\usepackage{caption}
\usepackage{subcaption}
\usetikzlibrary{pgfplots.groupplots}

\definecolor{confidence}{rgb}{0.12, 0.46, 0.71}
\definecolor{ece}{rgb}{0.84, 0.15, 0.16}

\newcommand{\realdigits}{\mathbb{R}}
\newcommand{\realdigitspositive}{\mathbb{R}_{>0}}

\newcommand{\pdf}{f}
\newcommand{\cdf}{F}
\newcommand{\ppf}{F^{-1}}
\newcommand{\prob}{\mathbb{P}}

\newcommand{\sampledfrom}{\sim}

\newcommand{\gp}{\text{gp}}
\newcommand{\kernel}{k}
\newcommand{\coregion}{\mathbf{B}}
\newcommand{\numinducing}{N^\ast}

\newcommand{\boundary}{a}

\newcommand{\normaldistribution}{\mathcal{N}}
\newcommand{\cauchydistribution}{Cauchy}

\newcommand{\expectation}{\mathbb{E}}

\newcommand{\mean}{\mu}
\newcommand{\meanvec}{\boldsymbol{\mu}}
\newcommand{\stddev}{\sigma}
\newcommand{\variance}{\sigma^2}
\newcommand{\cov}{\boldsymbol{\Sigma}}
\newcommand{\correlation}{\rho}

\newcommand{\decomposed}{\mathbf{L}}
\newcommand{\diagonal}{\mathbf{D}}

\newcommand{\quantile}{\tau}
\newcommand{\identity}{\mathbf{I}}

\newcommand{\cauchymode}{x_0}
\newcommand{\cauchyscale}{\gamma}

\newcommand{\T}{\top}

\newcommand{\inputvariate}{X}
\newcommand{\outputvariate}{Y}

\newcommand{\distvariate}{S}

\newcommand{\alloutputvariates}{\mathbf{\outputvariate}}

\newcommand{\inputset}{\mathcal{\inputvariate}}
\newcommand{\outputset}{\mathcal{Y}}

\newcommand{\distset}{\mathcal{\distvariate}}
\newcommand{\dataset}{\mathcal{D}}

\newcommand{\singleinput}{x}
\newcommand{\singleoutput}{y}

\newcommand{\dist}{s}

\newcommand{\numparams}{J}

\newcommand{\numdims}{K}

\newcommand{\numbins}{M}
\newcommand{\indexbins}{m}

\newcommand{\numsamples}{N}
\newcommand{\indexsamples}{i}

\newcommand{\parameter}{\theta}

\newcommand{\loss}{\mathcal{L}}

\newcommand{\nees}{\epsilon}

\newcommand{\model}{h}

\newcommand{\ind}{\mathds{1}}

\newcommand{\scaleweight}{w}
\newcommand{\scalevec}{\mathbf{w}}

\newcommand{\scaleset}{\mathcal{W}}

\newcommand{\diff}{\partial}

\newcommand{\fasterrcnn}[1][]{%
\ifthenelse{\equal{#1}{}}{\texttt{Faster R-CNN}}{\texttt{Faster R-CNN R#1-FPN}}%
}

\newcommand{\retinanet}[1][]{%
\ifthenelse{\equal{#1}{}}{\texttt{RetinaNet}}{\texttt{RetinaNet R#1-FPN}}%
}

\begin{document}
	\pagestyle{headings}
	\mainmatter
	\def\ECCVSubNumber{133}  
	
    \title{Parametric and Multivariate Uncertainty Calibration for Regression and Object Detection}

	\titlerunning{Parametric and Multivariate Uncertainty Calibration}
	%
	\author{Fabian Küppers\inst{1,2} \and
        Jonas Schneider\inst{1} \and
        Anselm Haselhoff\inst{2}}
    \authorrunning{F. Küppers et al.}
    %
    \institute{e:fs TechHub GmbH, Gaimersheim, Germany \\
        \email{\{fabian.kueppers,jonas.schneider\}@efs-auto.com} \vspace{0.5em} \and 
        Ruhr West University of Applied Sciences, Bottrop, Germany \\
        \email{anselm.haselhoff@hs-ruhrwest.de}
    }
	
	\maketitle
	
	\begin{abstract}
    Reliable spatial uncertainty evaluation of object detection models is of special interest and has been subject of recent work. In this work, we review the existing definitions for uncertainty calibration of probabilistic regression tasks. We inspect the calibration properties of common detection networks and extend state-of-the-art recalibration methods. Our methods use a Gaussian process (GP) recalibration scheme that yields parametric distributions as output (e.g. Gaussian or Cauchy). The usage of GP recalibration allows for a local (conditional) uncertainty calibration by capturing dependencies between neighboring samples. The use of parametric distributions such as as Gaussian allows for a simplified adaption of calibration in subsequent processes, e.g., for Kalman filtering in the scope of object tracking.
    
    In addition, we use the GP recalibration scheme to perform \textbf{covariance estimation} which allows for post-hoc introduction of local correlations between the output quantities, e.g., position, width, or height in object detection. To measure the joint calibration of multivariate and possibly correlated data, we introduce the \textit{quantile calibration error} which is based on the Mahalanobis distance between the predicted distribution and the ground truth to determine whether the ground truth is within a predicted quantile.
    
    Our experiments show that common detection models overestimate the spatial uncertainty in comparison to the observed error. We show that the simple Isotonic Regression recalibration method is sufficient to achieve a good uncertainty quantification in terms of calibrated quantiles. In contrast, if normal distributions are required for subsequent processes, our GP-Normal recalibration method yields the best results. Finally, we show that our \textbf{covariance estimation} method is able to achieve best calibration results for joint multivariate calibration.
    All code is open source and available at 
    \url{https://github.com/EFS-OpenSource/calibration-framework}.

\end{abstract}
	
	\clearpage
\begin{figure}[ht!]
	\centering
	\begin{subfigure}[t]{0.3\textwidth}
        \includegraphics[width=\linewidth]{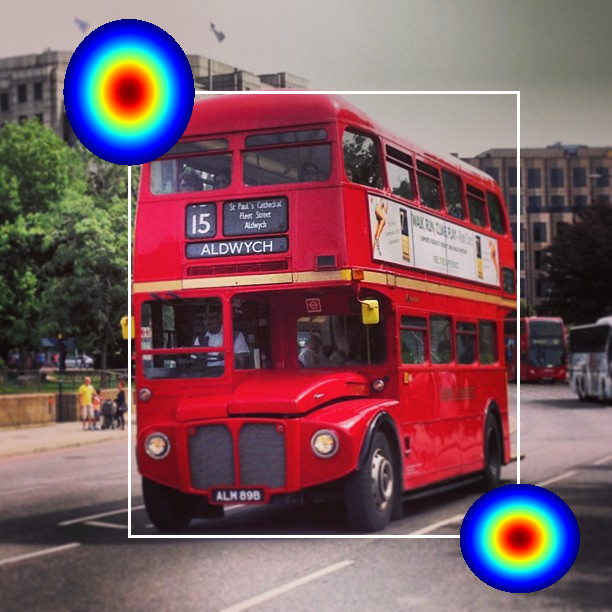}
    \end{subfigure}%
    \hfill%
    \begin{subfigure}[t]{0.3\textwidth}
        \includegraphics[width=\linewidth]{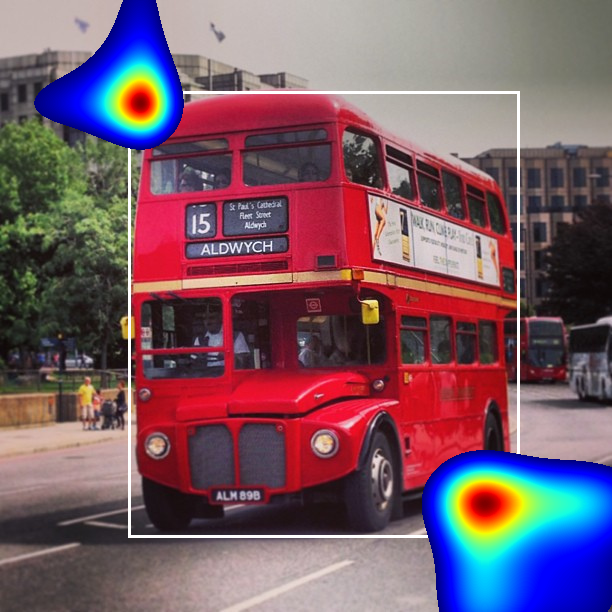}
    \end{subfigure}%
    \hfill%
    \begin{subfigure}[t]{0.3\textwidth}
        \includegraphics[width=\linewidth]{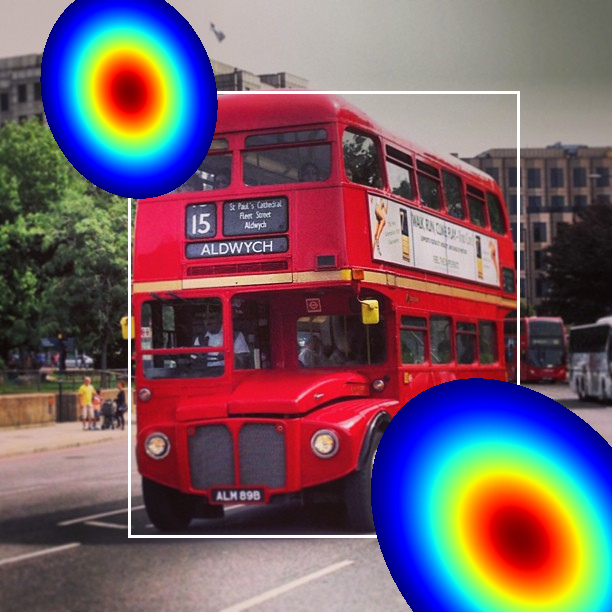}
    \end{subfigure}%
	\caption{Qualitative example of spatial uncertainty regression for object detection tasks. (a) A probabilistic \retinanet{}\cite{Lin2017} outputs (independent) normal distributions for each dimension. (b) On the one hand, we can recalibrate these distributions using \textit{GP-Beta} \cite{Song2019} which yields multiple (independent) distributions of arbitrary shape. (c) On the other hand, we can also use our multivariate (mv.) \textit{GP-Normal} method which recalibrates the uncertainty of Gaussian distributions and is also able to capture correlations between different dimensions.}
	\label{fig:qualitative}
\end{figure}

\section{Introduction}

Obtaining reliable uncertainty information is a major concern especially for safety-critical applications such as autonomous driving \cite{Yang2018,Feng2019,Feng2021}.
For camera-based environment perception, it is nowadays possible to utilize object detection algorithms that are based on neural networks \cite{Banerjee2018}.
Such detection methods can be used in the context of object tracking (tracking-by-detection) to track individual objects within an image sequence \cite{Bar2004,Van2005}.
Besides the position of individual objects, common tracking algorithms such as Kalman filtering require additional uncertainty information.
However, these detection models are known to produce unreliable confidence information \cite{Guo2018,Kueppers2020,Schwaiger2021}.
In this context, uncertainty calibration is the task of matching the predicted model uncertainty with the observed error. For example, in the scope of classification, calibration requires that the predicted confidence of a forecaster should match the observed accuracy. 
In contrast, the task for probabilistic regression models is to target the true ground-truth score for an input using a probability distribution as the model output \cite{Kendall2017,He2019}.
An object detection model can also be trained to output probabilistic estimates for the position information using Gaussian distributions \cite{He2019,Hall2020,Harakeh2020,Feng2021}.
We refer to this as the probabilistic regression branch of a detection model.

While there is a large consent about the metrics and definitions in the scope of classification calibration \cite{Naeini2015,Naeini2015b,Guo2018,Kull2017,Kumar2019}, research for regression calibration still differs in definitions as well as in the used evaluation metrics \cite{Kuleshov2018,Song2019,Levi2019,Laves2020}.
Recent work has proposed several methods to apply post-hoc calibration for regression tasks \cite{Kuleshov2018,Levi2019,Song2019,Laves2020}.
A flexible and input-dependent recalibration method is the GP-Beta provided by \cite{Song2019} that utilizes a Gaussian Process (GP) for recalibration parameter estimation.
However, this method provides non-parametric distributions as calibration output which might be disadvantageous for subsequent applications such as Kalman filtering since these applications commonly require parametric distributions, e.g., Gaussians \cite{Bar2004}.
In contrast, the authors in \cite{Levi2019,Laves2020} propose a temperature scaling \cite{Guo2018} for the variance of a Gaussian distribution which, however, is not sensitive to a specific input.
Furthermore, most of the current research only focuses on the calibration of 1D regression problems. However, especially for object detection, it might be necessary to jointly inspect the calibration properties in all dimensions of the probabilistic regression branch.
For example, a large object might require a different uncertainty calibration compared to smaller objects. A representative uncertainty quantification is of special interest especially for safety-critical applications such as autonomous driving or tracking tasks, e.g., Kalman filtering. However, a joint recalibration of multiple dimensions is not possible using the current existing regression calibration methods.
Therefore, we seek for a multivariate regression recalibration method that offers the flexibility in the parameter estimation of methods such as GP-Beta \cite{Song2019} but also provides parametric probability distributions.\\

\textbf{Contributions.} In this work, we focus on the safety-relevant task of object detection and provide a brief overview over the most important definitions for regression calibration. We adapt the idea of Gaussian process (GP) recalibration \cite{Song2019} by using parametric probability distributions (Gaussian or Cauchy) as calibration output. Furthermore, we investigate a method for a joint regression calibration of multiple dimensions. The effect of joint uncertainty calibration is qualitatively demonstrated in Fig.~\ref{fig:qualitative}. On the one hand, this method is able to capture possible (conditional) correlations within the data to subsequently learn covariances. On the other hand, our method is also able to recalibrate multivariate Gaussian probability distributions with full covariance matrices. The task of joint multivariate recalibration for regression problems hasn't been addressed so far to best of our knowledge.
	\section{Definitions for Regression Calibration \& Related Work}
In this section, we give an overview over the related work regarding uncertainty calibration for regression. Furthermore, we summarize the most important definitions for regression calibration. 
In this way, we can examine calibration in a unified context that allows us to better compare the calibration definitions as well as the associated calibration methods.

In the scope of uncertainty calibration, we work with models that output an uncertainty quantification for each forecast. 
Let $\inputvariate \in \inputset$ denote the input for a regression model $\model(\inputvariate)$ that predicts the aleatoric (data) uncertainty as a probability density distribution $\pdf_{\outputvariate|\inputvariate} \in \distset$ where $\distset$ denotes the set of all possible probability distributions, so that $\distvariate_\outputvariate := \pdf_{\outputvariate|\inputvariate}(\singleoutput | \singleinput) = \model(\singleinput)$, $\distvariate_\outputvariate \in \mathcal{S}$. The predicted probability distribution targets the ground-truth $\outputvariate \in \outputset = \realdigits$ in the output space. Let further denote $\cdf_{\outputvariate|\inputvariate}: \outputset \rightarrow (0, 1)$ the respective cumulative density function (CDF), and $\ppf_{\outputvariate|\inputvariate}: (0, 1) \rightarrow \outputset$ the (inverse) quantile function.

\textbf{Quantile calibration.} We start with the first definition for regression calibration in terms of quantiles provided by \cite{Kuleshov2018}. The authors argue that a probabilistic forecaster is well calibrated if an estimated prediction interval for a certain quantile level $\quantile \in (0, 1)$ covers the ground-truth $\outputvariate$ in $100\quantile\%$ cases, i.e., if the predicted and the empirical CDF match given sufficient data. 
More formally, a probabilistic forecaster is \textit{quantile calibrated}, if
\begin{align}
	\label{eq:quantile_calibration}
	\prob\Big(\outputvariate \leq \ppf_{\outputvariate|\inputvariate}(\quantile)\Big) = \quantile, \quad \forall \quantile \in (0, 1) ,
\end{align}
is fulfilled. This also holds for two-sided quantiles \cite{Kuleshov2018}. Besides their definition of \textit{quantile calibration}, the authors in \cite{Kuleshov2018} propose to use the isotonic regression calibration method \cite{Zadrozny2002} from classification calibration to recalibrate the cumulative distribution function predicted by a probabilistic forecaster. We adapt this method and use it as a reference in our experiments.\\

\textbf{Distribution calibration.} In classification calibration, the miscalibration of a classifier is conditioned on the predicted confidence \cite{Guo2018}, i.e., we inspect the miscalibration of samples with equal confidence. In contrast, the definition of \textit{quantile calibration} in (\ref{eq:quantile_calibration}) only faces the marginal probability for all distributional moments \cite{Song2019,Laves2020,Levi2019}. 
Therefore, the authors in \cite{Song2019} extend the definition for regression calibration to \textit{distribution calibration}. This definition requires that an estimated probability distribution should match the observed error distribution given sufficient data. 
Therefore, a model is \textit{distribution calibrated}, if 
\begin{align}
	\label{eq:distribution_calibration}
	\pdf(\outputvariate = \singleoutput | \distvariate_\outputvariate = \dist) = \dist(\singleoutput)
\end{align}
holds for all continuous distributions $\dist \in \distset$ that target $\outputvariate$, and $\singleoutput \in \outputset$ \cite{Song2019}. Following Theorem 1 in \cite{Song2019}, a \textit{distribution calibrated} probabilistic forecaster is also \textit{quantile calibrated} (but not vice-versa).
In addition, the authors in \cite{Song2019} propose the \textit{GP-Beta} recalibration method, an approach that uses beta calibration \cite{Kull2017} in conjunction with GP parameter estimation to recalibrate the cumulative distribution and to achieve \textit{distribution calibration}. The \textit{Isotonic Regression} as well as the \textit{GP-Beta} approaches exploit the fact that the CDF is defined in the $[0, 1]$ interval which allows for an application of recalibration methods from the scope of classification calibration. While the former method \cite{Kuleshov2018} seeks for the optimal recalibration parameters globally, the latter approach \cite{Song2019} utilizes GP parameter estimation to find the optimal recalibration parameters for a single distribution using all samples whose distribution moments are close to the target sample. This allows for a local uncertainty recalibration.\\

\textbf{Variance calibration.} Besides the previous definitions, the authors in \cite{Levi2019} and \cite{Laves2020} consider normal distributions over the output space with mean $\mean_\outputvariate(\inputvariate)$ and variance $\variance_\outputvariate(\inputvariate)$ that are implicit functions of the base network with $\inputvariate$ as input. The authors define the term of \textit{variance calibration}. This definition requires that the predicted variance of a forecaster should match the observed variance \textit{for a certain variance level}. For example, given a certain set of predictions by an unbiased probabilistic forecaster \textit{with equal variance}, the mean squared error (which is equivalent to the observed variance in this case) should match the predicted variance. This must hold for all possible variances. Thus, \textit{variance calibration} is defined as
\begin{align}
	\expectation_{\inputvariate, \outputvariate} \Big[ \Big(\outputvariate - \mean_\outputvariate(\inputvariate) \Big)^2 \Big| \variance_\outputvariate(\inputvariate) = \variance \ \Big] = \variance ,
\end{align}
for all $\variance \in \realdigitspositive$ \cite{Levi2019,Laves2020}. Note that any \textit{variance calibrated} forecaster is \textit{quantile calibrated}, if (a) the forecaster is unbiased, i.e., $\expectation_{\outputvariate|\inputvariate}[\outputvariate] = \mean = \mean_\outputvariate(\inputvariate)$ for all $\inputvariate$, and (b) the ground-truth data is normally distributed around the predicted mean. In this case, the predicted normal distribution $\normaldistribution(\mean_\outputvariate(\inputvariate), \variance_\outputvariate(\inputvariate))$ matches the observed data distribution $\normaldistribution(\mean, \variance)$ for any $\variance \in \realdigitspositive$, and thus $\ppf_{\outputvariate|\inputvariate}(\quantile) = \ppf_\text{obs}(\quantile)$ for any $\quantile \in (0, 1)$. If the observed variance does not depend on the mean, i.e., $\text{Cov}(\mean, \variance) = 0$, and conditions (a) and (b) are fulfilled, then a \textit{variance calibrated} forecaster is also \textit{distribution calibrated}, as the condition in (\ref{eq:distribution_calibration}) is met because $\distset$ is restricted to normal distributions and $\mean$ is no influential factor.
For uncertainty calibration, the authors in \cite{Levi2019} and \cite{Laves2020} use temperature scaling \cite{Guo2018} to rescale the variance of a normal distribution. This method optimizes the negative log likelihood (NLL) to find the optimal rescaling parameter. \\

In conclusion, although the definition of \textit{quantile calibration} \cite{Kuleshov2018} is straightforward and intuitive, only the marginal probabilities are considered which are independent of the predicted distribution moments. In contrast, the definition for \textit{distribution calibration} \cite{Song2019} is more restrictive as it requires a forecaster to match the observed distribution conditioned on the forecaster's output. If a forecaster is distribution calibrated and unbiased, and if the ground-truth data follows a normal distribution, then the definition for \textit{variance calibration} \cite{Levi2019,Laves2020} is met which is the most restrictive definition for regression calibration but also very useful when working with normal distributions. \\

\textbf{Further research.} A different branch of yielding prediction uncertainties is quantile regression where a forecaster outputs estimates for certain quantile levels \cite{Fasiolo2020,Chung2021}. The advantage of quantile regression is that no parametric assumptions about the output distribution is necessary. However, these methods haven't been used for object detection so far, thus, we focus on detection models that output a parametric estimate for the object position \cite{He2019,Hall2020,Harakeh2020,Feng2021}.
As opposed to post-hoc calibration, there are also methods that aim to achieve regression calibration during model training, e.g., calibration loss \cite{Feng2019}, maximum mean discrepancy (MMD) \cite{Cui2020} or $f$-Cal \cite{Bhatt2021}. The former approach adds a loss term to match the predicted variances with the observed squared error, while the latter methods introduce a second loss term to perform distribution matching during model training. The advantage of these approaches is that an additional calibration data set is not needed. However, it is necessary to retrain the whole network architecture. Thus, we focus on post-hoc calibration methods in this work.
	\section{Joint Parametric Regression Calibration}
Several tasks such as object detection are inference problems with multiple dimensions, i.e., the joint estimation of position, width, and height of objects is necessary. Therefore, the output space extends to $\outputset = \realdigits^{\numdims}$, where $\numdims$ denotes the number of dimensions. In the scope of object detection, most probabilistic models predict independent distributions for each quantity \cite{He2019,Hall2020,Feng2021}.

We start with the GP-Beta framework by \cite{Song2019}. Given a training set $\dataset$, the goal is to build a calibration mapping from uncalibrated distributions to calibrated ones. Since the CDF of a distribution is bound to the $[0,1]$ interval, the authors in \cite{Song2019} derive a \textit{beta link function} from the beta calibration method \cite{Kull2017} to rescale the CDF of a distribution with a total amount of 3 rescaling parameters. 
These rescaling parameters $\scalevec \in \scaleset = \realdigits^3$ are implicitly obtained by a GP model $\gp(0, \kernel, \coregion)$ with zero mean, where $\kernel$ is the covariance kernel function of the GP, and $\coregion$ is the coregionalization matrix that captures correlations between the parameters \cite{Song2019}. 
The authors argue that the usage of a GP model allows for finding the optimal recalibration parameters w.r.t. distributions that are close to the actual prediction.
During model training, the parameters $\scalevec$ for the beta link function are trained using the training data distribution given by
\begin{align}
    \label{eq:training:likelihood}
    \pdf(\outputvariate | \dataset) = \int_\scaleset \pdf (\outputvariate | \scalevec, \dataset) \pdf (\scalevec | \dataset) \diff \scalevec ,
\end{align}
with the likelihood $\pdf (\outputvariate | \scalevec, \dataset)$ obtained by the beta link function and $\pdf (\scalevec | \dataset)$ as the posterior obtained by the GP model.
However, the integral in (\ref{eq:training:likelihood}) is analytically intractable due to the non-linearity within the link function in the likelihood \cite{Song2019}.
Thus, the authors in \cite{Song2019} use a Monte-Carlo sampling approach for model training and inference.
Furthermore, the authors use an approximate GP because the GP covariance matrix has the complexity $\mathcal{O}(\numsamples^2)$. Thus, during optimization, $\numinducing$ pseudo inducing points are learnt to represent the training set \cite{Snelson2006,Hensman2015}.
This guarantees a stable and fast computation of new calibrated distributions during inference.
The inducing points, the coregionalization matrix, and the kernel parameters are optimized using the evidence lower bound (ELBO). We refer the reader to \cite{Song2019} for a detailed derivation of the beta link likelihood.\\

\begin{figure}[t!]
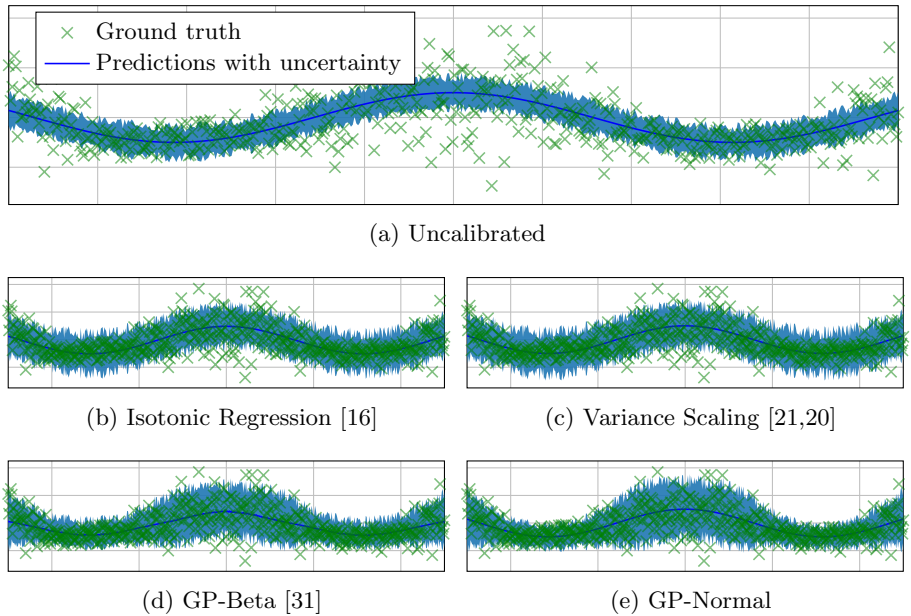

	\centering
	\begin{subfigure}[t]{0.99\linewidth}
		\input{img/artificial/artificial_0.tikz}
		\subcaption{Uncalibrated}
	\end{subfigure}
	\vspace{1em}
	
	\begin{subfigure}[t]{0.5\textwidth}
		\input{img/artificial/artificial_1.tikz}
		\subcaption{Isotonic Regression \cite{Kuleshov2018}}
	\end{subfigure}%
	\begin{subfigure}[t]{0.5\textwidth}
		\input{img/artificial/artificial_2.tikz}
		\subcaption{Variance Scaling \cite{Levi2019,Laves2020}}
	\end{subfigure}%
	\vspace{1em}
	
	\begin{subfigure}[t]{0.5\textwidth}
		\input{img/artificial/artificial_3.tikz}
		\subcaption{GP-Beta \cite{Song2019}}
	\end{subfigure}%
	\begin{subfigure}[t]{0.5\textwidth}
		\input{img/artificial/artificial_4.tikz}
		\subcaption{GP-Normal}
	\end{subfigure}%
	\caption{Regression example using a cosine function with noise to generate ground-truth samples (green). The noise increases towards the function's maximum. (a) shows an unbiased estimator with equally sampled variance (blue). (b) \textit{Isotonic Regression} accounts for marginal uncertainty across all samples, while (c) \textit{Variance Scaling} only has a dependency on the predicted variance. 
	In contrast, (d) \textit{GP-Beta} and (e) \textit{GP-Normal} capture information from the whole predicted distribution and thus are able to properly recalibrate the uncertainty.}
	\label{fig:artificial}
\end{figure}

\textbf{Parametric recalibration.} A major advantage of this approach is that it captures dependencies between neighboring input samples,
i.e., it reflects a conditional recalibration of the estimated probability distributions. However, the resulting distribution has no parametric form as it is represented by a CDF of arbitrary shape. For this reason, we propose a novel recalibration method \textbf{GP-Normal} which can be seen as a temperature scaling that rescales the variance $\variance$ by a scaling factor $\scaleweight$ \cite{Laves2020,Levi2019}. In contrast to the baseline approaches in \cite{Laves2020,Levi2019}, we use the GP scheme by \cite{Song2019} to estimate a weight $\scaleweight \in \realdigitspositive$ for each distribution individually, so that
\begin{align}
    \label{eq:scaleweight:univariate}
	\log(\scaleweight) \sampledfrom \gp(0, \kernel) 
\end{align}
can be used to rescale the variance $\hat{\stddev}^2 = \scaleweight \cdot \variance$ with the same training likelihood given in (\ref{eq:training:likelihood}).
Note that a coregionalization matrix $\coregion$ is not required as only a single parameter is estimated. We use the exponential function to guarantee positive estimates for $\scaleweight$. The advantage of this approach compared to the standard \textit{GP-Beta} is that the output distribution has a closed-form representation. Furthermore, the uncertainty recalibration is sensitive to neighboring samples, i.e., it allows for a conditional uncertainty recalibration of the input distributions, compared to the standard \textit{Isotonic Regression} \cite{Kuleshov2018} or \textit{Variance Scaling} \cite{Laves2020,Levi2019}. 
A comparison of common calibration methods and our \textit{GP-Normal} method is given in Fig.~\ref{fig:artificial} for an artificial example.

\begin{figure}[t!]
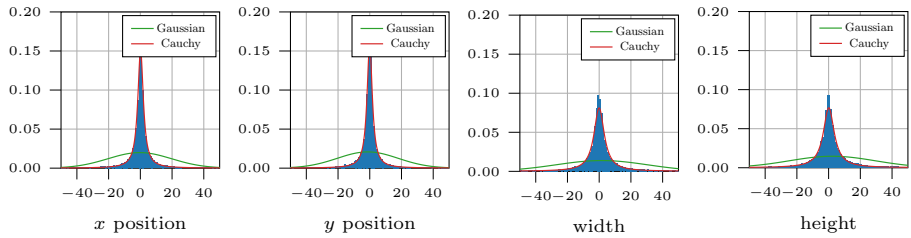

	\centering
	\begin{subfigure}[t]{0.25\textwidth}
		\input{img/error_distribution/error_histogram_dim_0.tikz}
	\end{subfigure}%
	\begin{subfigure}[t]{0.25\textwidth}
		\input{img/error_distribution/error_histogram_dim_1.tikz}
	\end{subfigure}%
	\begin{subfigure}[t]{0.25\textwidth}
		\input{img/error_distribution/error_histogram_dim_2.tikz}
	\end{subfigure}%
	\begin{subfigure}[t]{0.25\textwidth}
		\input{img/error_distribution/error_histogram_dim_3.tikz}
	\end{subfigure}
	\vspace{-1em}
	\caption{Prediction error of the mean estimate $\meanvec_\alloutputvariates(\inputvariate)$ of a \fasterrcnn on MS COCO. We can only get a poor quality of fit using a Gaussian distribution in this case. In contrast, the Cauchy distribution achieves a good approximation of the error distribution as this distribution allows for heavier tails.}
	\label{fig:errordist}
\end{figure}

In general, we can use any parametric continuous probability distribution to derive a calibrated distribution. For example, if we inspect the distribution of the error obtained by a \fasterrcnn \cite{Ren2015} on the MS COCO dataset \cite{Lin2014} (shown in Fig.~\ref{fig:errordist}), we can observe that the error rather follows a Cauchy distribution $\cauchydistribution(\cauchymode, \cauchyscale)$ with location parameter $\cauchymode \in \realdigits$ and scale $\cauchyscale \in \realdigitspositive$. Therefore, we further propose the \textbf{GP-Cauchy} method that is similar to \textit{GP-Normal} but utilizes a Cauchy distribution as the likelihood within the GP framework. This allows for deriving a scale weight $\scaleweight$ (cf. (\ref{eq:scaleweight:univariate})) so that $\hat{\cauchyscale} = \scaleweight \cdot \stddev$ leads to a calibrated probability distribution $\cauchydistribution(\cauchymode, \hat{\cauchyscale})$ using $\mean_\outputvariate(\inputvariate) = \cauchymode$ as an approximation. \\

\textbf{Joint recalibration.} In many applications, such as object detection, a probabilistic regression model needs to jointly infer the output distribution for multiple dimensions. Usually, the output of such a model is parameterized using independent Gaussian distributions for each dimension \cite{Kendall2017,He2019,Hall2020}. We can adapt the \textit{GP-Normal}, \textit{GP-Cauchy}, as well as the \textit{GP-Beta} methods \cite{Song2019} to jointly recalibrate these independent distributions.
Since we assert normal distributions as calibration input, we use the same kernel function as \cite{Song2019} and developed by \cite{Song2008}, that is defined by
\begin{align}
	\kernel\big((\meanvec_i, \cov_i), (\meanvec_j, \cov_j)\big) = \parameter^\numdims | \cov_{ij} |^{-\frac{1}{2}} \exp\Big(-\frac{1}{2}(\meanvec_i - \meanvec_j)^\T \cov_{ij}^{-1}(\meanvec_i - \meanvec_j)\Big) ,
\end{align}
where $\parameter \in \realdigits$ is the kernel lengthscale parameter and $\cov_{ij} = \cov_i + \cov_j + \parameter^2\identity$ defines the covariance, with $\identity$ as the identity matrix.
In the \textit{GP-Beta} framework, the coregionalization matrix $\coregion$ is used to introduce correlations in the output data. We extend this framework to \textit{jointly} recalibrate probability distributions within a multidimensional output space $\alloutputvariates \in \outputset = \realdigits^\numdims$.
We extend the coregionalization matrix $\coregion$ by additional entries for each dimension. Therefore, the parameter estimation in (\ref{eq:scaleweight:univariate}) for the \textit{GP-Normal} and the \textit{GP-Cauchy} methods extends to
\begin{align}
	\log(\scalevec) \sampledfrom \gp(0, \kernel, \coregion) ,
\end{align}
where $\coregion \in \realdigits^{\numdims \times \numdims}$ and $\scalevec \in \realdigitspositive^\numdims$. 
The likelihood of GP calibration methods can be interpreted as the product of $\numdims$ multiple independent dimensions.\\

\textbf{Covariance estimation.} Since the \textit{GP-Normal} method yields a parametric normal distribution, it is also possible to capture correlations between all output quantities in $\outputset$ by introducing a \textbf{covariance estimation} scheme. First, we capture the \textit{marginal} correlation coefficients $\correlation_{ij} \in [-1, 1]$ between dimensions $i$ and $j$. This allows for the computation of a covariance matrix $\cov$ for all samples in the training set $\dataset$. Second, we use the $\decomposed\diagonal\decomposed^\T$ factorization of $\cov$ and perform a rescaling of the lower triangular $\decomposed$ and the diagonal matrix $\diagonal$ by the weights $\scalevec_\decomposed \in \realdigits^{\numparams-\numdims}$ and $\scalevec_\diagonal \in \realdigitspositive^\numdims$, respectively, with the total number of parameters $\numparams = \frac{K}{2}(K+1) + \numdims$, where 
\begin{align}
	\log(\scalevec_\diagonal), \scalevec_\decomposed \sampledfrom \gp(0, k, \coregion),
\end{align}
using $\coregion \in \realdigits^{\numparams \times \numparams}$. After covariance reconstruction, the likelihood of the model is obtained by the multivariate normal distribution using the mean $\meanvec$ and the rescaled covariance matrix $\hat{\cov} = \hat{\decomposed}\hat{\diagonal}\hat{\decomposed}^\T$, where $\hat{\decomposed} = \scalevec_\decomposed \odot \decomposed$ and $\hat{\diagonal} = \scalevec_\diagonal \odot \diagonal$ using the broadcasted weights ($\odot$ denotes the element-wise multiplication).
On the one hand, it is now possible to introduce conditional correlations between independent Gaussian distributions using the \textit{GP-Normal} 
method. On the other hand, we can also use this method for \textbf{covariance recalibration} if covariance estimates are already given in the input data set $\dataset$ \cite{Harakeh2020}. In this case, we can directly use the $\decomposed\diagonal\decomposed^\T$ factorization on the given covariance matrices to perform a covariance recalibration. Both approaches, \textit{covariance estimation} and \textit{covariance recalibration}, might be used to further improve the uncertainty quantification. This is beneficial if these uncertainties are incorporated in subsequent processes. 

\section{Measuring Miscalibration}
Besides proper scoring rules such as negative log likelihood (NLL), there is actually no common consent about how to measure regression miscalibration. A common metric to measure the quality of predicted quantiles is the \textit{Pinball loss} $\loss_{\text{Pin}}(\quantile)$ \cite{Steinwart2011} 
for a certain quantile level $\quantile$. For regression calibration evaluation, a mean Pinball loss $\overline{\loss_\text{Pin}} = \expectation_\quantile[\loss_\text{Pin}(\tau)]$ for multiple quantile levels is commonly used to get an overall measure. Recently, the authors in \cite{Laves2020} defined the \textit{uncertainty calibration error} (UCE) in the scope of \textit{variance calibration} that measures the difference between predicted uncertainty and the actual error. Similar to the \textit{expected calibration error} (ECE) within classification calibration, the UCE uses a binning scheme with $\numbins$ bins over the variance to estimate the calibration error w.r.t. the predicted variance. More formally, the UCE is defined by
\begin{align}
	\text{UCE} := \sum^{\numbins}_{\indexbins=1} \frac{\numsamples_\indexbins}{\numsamples} | \text{MSE}(\indexbins) - \text{MV}(\indexbins) | ,
\end{align}
with $\text{MSE}(\indexbins)$ and $\text{MV}(\indexbins)$ as the mean squared error and the mean variance within each bin, respectively, and $\numsamples_\indexbins$ as the number of samples within bin $\indexbins$ \cite{Laves2020}. Another similar metric, the \textit{expected normalized calibration error} (ENCE), has been proposed by \cite{Levi2019} and is defined by
\begin{align}
	\text{ENCE} := \frac{1}{\numbins} \sum^{\numbins}_{\indexbins=1} \frac{| \text{RMSE}(\indexbins) - \text{RMV}(\indexbins) |}{\text{RMV}(\indexbins)} ,
\end{align}
where $\text{RMSE}(\indexbins)$ and $\text{RMV}(\indexbins)$ denote the root mean squared error and the root mean variance within each bin, respectively. The advantage of the ENCE is that the miscalibration can be expressed as the percentage of the $\text{RMV}$. Thus, the error is independent of the output space.\\

\textbf{Measuring multivariate miscalibration.} Besides NLL, none of these metrics is able to jointly measure the miscalibration of multiple dimensions or to capture the influence of correlations between multiple dimensions. Therefore, we derive a new metric, the \textbf{quantile calibration error} (QCE). This metric is based on the well-known \textit{normalized estimation error squared} (NEES) that is used for Kalman filter consistency evaluation \cite[pp.~232]{Bar2004}, \cite[pp.~292]{Van2005}. Given an unbiased base estimator that outputs a mean $\meanvec_\alloutputvariates(\inputvariate)$ and covariance $\cov_\outputvariate(\inputvariate)$ estimate with $\numdims$ dimensions for an input $\inputvariate$, the NEES $\nees$ is the squared Mahalanobis distance
between the predicted distribution $\pdf_\outputvariate$ and the ground-truth $\mathbf{\outputvariate}$. The hypothesis, that the predicted estimation errors are consistent with the observed estimation errors, is evaluated using a $\chi^2$-test. This test is accepted if the mean NEES is below a certain quantile threshold $\boundary_\quantile$ for a target quantile $\quantile$, i.e., $\nees_{\meanvec, \cov} \leq \boundary_\quantile$. The threshold $\boundary_\quantile \in \realdigitspositive$ is obtained by a $\chi^2_\numdims$ distribution with $\numdims$ degrees of freedom, so that $\boundary_\quantile = \chi^2_\numdims(\quantile)$. The advantage of the NEES is that it measures a \textit{normalized} estimation error and can also capture correlations between multiple dimensions.

For regression calibration evaluation, we adapt this idea and measure the difference between the fraction of samples that fall into the acceptance interval for a certain quantile $\quantile$ and the quantile level $\quantile$ itself. However, this would only reflect the \textit{marginal} calibration error. Therefore, we would like to measure the error conditioned on distributions with similar properties. For instance, the dispersion of a distribution can be captured using the \textit{standardized generalized variance} (SGV) which is defined by $\variance_G = \det(\cov)^{\frac{1}{\numdims}}$ \cite{Sengupta1987,Sengupta2004}. In the univariate case, the SGV is equal to the variance $\variance$. Using the SGV as a property of the distributions, we define the QCE as
\begin{align}
	\text{QCE}(\quantile) := \expectation_{\inputvariate, \stddev_G} \Big[ \big| \prob( \nees(\inputvariate) \leq \boundary_\quantile | \stddev_G) - \quantile \big| \Big] .
\end{align}
Similar to the ECE, UCE, and ENCE, we use a binning scheme but rather over the square root $\stddev_G$ of the SGV to achieve a better data distribution for binning. The QCE measures the fraction of samples whose NEES falls into the acceptance interval for each bin, separately. Therefore, the QCE for a certain quantile $\quantile$ is approximated by
\begin{align}
	\text{QCE}(\quantile) \approx \sum^{\numbins}_{\indexbins=1} \frac{\numsamples_\indexbins}{\numsamples} \Big| \text{freq}(\indexbins) - \quantile \Big| ,
\end{align}
where $\text{freq}(\indexbins) = \frac{1}{\numsamples_\indexbins} \sum^{\numsamples_\indexbins}_{\indexsamples=1} \ind\big(\nees(\singleinput_\indexsamples) \leq \boundary_\quantile\big)$
using $\ind(\cdot)$ as the indicator function.
Similar to the Pinball loss, we can approximate a mean $\overline{\text{QCE}} = \expectation_\quantile[\text{QCE}(\quantile)]$ for multiple quantile levels to get an overall quality measure. Furthermore, we can use this metric to measure the quality of multivariate uncertainty estimates in the following experiments. Note that we can also use the QCE to measure non-Gaussian distributed univariate random variables. The derivation of the QCE for multivariate, non-parametric distributions is subject of future work.
	\section{Experiments}

We evaluate our methods using probabilistic object detection models that are trained on the Berkeley DeepDrive 100k \cite{Yu2018} and the MS COCO 2017 \cite{Lin2014} data sets.
We use a \retinanet[50]{} \cite{Lin2017} and a \fasterrcnn[101]{} \cite{Ren2015} as base networks. These networks are trained using a probabilistic regression output where a mean and a variance score is inferred for each bounding box quantity \cite{Kendall2017,He2019,Hall2020}. For uncertainty calibration training and evaluation, these networks are used to infer objects on the validation sets of the Berkeley DeepDrive 100k and COCO 2017 data sets.
Unfortunately, we cannot use the respective test sets as we need the ground-truth information for both, calibration training and evaluation. Therefore, we split both validation datasets and use the first half for the training of the calibration methods, while the second half is used for calibration evaluation. Finally, it is necessary to match all bounding box predictions with the ground-truth objects to determine the estimation error. For all experiments, we use an intersection over union (IoU) score of $0.5$ for object matching.

For calibration training and evaluation, we adapt the methods \textit{Isotonic Regression} \cite{Kuleshov2018}, \textit{Variance Scaling} \cite{Levi2019,Laves2020}, and \textit{GP-Beta} \cite{Song2019} and provide an optimized reimplementation in our publicly available repository\footnote{\url{https://github.com/EFS-OpenSource/calibration-framework}}.
Furthermore, this repository also contains the new \textit{GP-Normal} method and its multivariate extension for covariance estimation and covariance recalibration, as well as the \textit{GP-Cauchy} method. The GP methods are based on GPyTorch \cite{Gardner2018} that provides an efficient implementation for GP optimization. For calibration evaluation, we use the negative log likelihood (NLL), Pinball loss, UCE \cite{Laves2020}, ENCE \cite{Levi2019}, and our proposed QCE. The UCE, ENCE, and QCE use binning schemes with $\numbins=20$ bins, respectively, while the Pinball loss and the QCE use confidence levels from $\quantile=0.05$ up to $\quantile=0.95$ within steps of $0.05$ to obtain an average calibration score over different quantile levels. Note that we cannot report the UCE and ENCE for the \textit{GP-Cauchy} method as the Cauchy distribution has no variance defined. Furthermore, we only report the multivariate QCE for Gaussian distributions in our experiments since this metric is based on the squared Mahalanobis distance for Gaussian distributions in the multivariate case. On the one hand, we examine the calibration properties of each dimension independently and denote the average calibration metrics over all dimensions. On the other hand, we further compute the multivariate NLL and QCE to measure calibration for multivariate distributions. The results are given in Table~\ref{tab:evaluation:univariate}.
\begin{table}[t!]
	\centering
	\caption{Calibration results for \fasterrcnn[101]{} and \retinanet[50]{} evaluated on Berkeley DeepDrive and MS COCO evaluation datasets before and after uncertainty calibration with different calibration methods.}
	\label{tab:evaluation:univariate}
	\begin{tabular}{l|c|l||c|c|c|c|c||c|c}
	   \multicolumn{3}{c||}{Setup} & \multicolumn{5}{c||}{Univariate} & \multicolumn{2}{c}{Multivariate}\\ \hline
		 & DB & Method & NLL & $\overline{\text{QCE}}$ & $\overline{\loss_\text{Pin}}$ & UCE & ENCE & NLL & $\overline{\text{QCE}}$ \\ \hline \hline
		 \multirow{14}{*}{\rotatebox[origin=c]{90}{\fasterrcnn{}}} & \multirow{7}{*}{BDD} & Uncalibrated & 3.053 & 0.040 & 1.079 & 19.683 & 0.454 & 12.210 & \textbf{0.071} \\
		 &  & Isotonic Reg. \cite{Kuleshov2018} & \textbf{2.895} & \textbf{0.017} & \textbf{1.059} & 39.157 & 0.303 & \textbf{11.579} & - \\
         &  & GP-Beta \cite{Song2019} & 2.941 & 0.057 & 1.077 & 3.635 & 0.199 & 11.764 & - \\
         &  & Var. Scaling \cite{Laves2020,Levi2019} & 2.962 & 0.061 & 1.086 & 3.361 & \textbf{0.175} & 11.848 & 0.131\\
         &  & GP-Normal & 2.962 & 0.059 & 1.084 & 3.289 & 0.188  & 11.848 & 0.128\\
         &  & GP-Normal (mv.) & 2.968 & 0.054 & 1.150 & \textbf{3.234} & 0.191 & 11.584 & 0.133 \\
         &  & GP-Cauchy & 3.011 & 0.050 & 1.189 & - & - & 12.045 & - \\ \cline{2-10}
		 & \multirow{7}{*}{COCO} & Uncalibrated & 3.561 & 0.154 & 3.055 & 32.899 & 0.096 & 14.245 & 0.256 \\
		 &  & Isotonic Reg. \cite{Kuleshov2018} & \textbf{3.340} & \textbf{0.020} & \textbf{2.715} & 42.440 & 0.121 & \textbf{13.360} & - \\
 		 &  & GP-Beta \cite{Song2019} & 3.412 & 0.074 & 2.750 & 51.455 & 0.140 & 13.649 & - \\
 		 &  & Var. Scaling \cite{Laves2020,Levi2019} & 3.554 & 0.131 & 2.952 & 33.155 & 0.093 & 14.216 & 0.222\\
		 &  & GP-Normal & 3.554 & 0.130 & 2.949 & 48.167 & 0.132 & 14.235 & \textbf{0.200}\\
		 &  & GP-Normal (mv.) & 3.562 & 0.121 & 3.298 & \textbf{28.382} & \textbf{0.087} & 13.955 & 0.216\\ 
		 &  & GP-Cauchy & 3.406 & 0.039 & 2.897 & - & - & 13.624 & - \\ \hline \hline
	     \multirow{14}{*}{\rotatebox[origin=c]{90}{\retinanet{}}} & \multirow{7}{*}{BDD} & Uncalibrated & 4.052 & 0.130 & 1.847 & 39.933 & 0.491 & 16.208 & 0.234 \\
         & & Isotonic Reg. \cite{Kuleshov2018} & \textbf{3.224} & \textbf{0.068} & \textbf{1.814} & 39.498 & 0.252 & \textbf{12.898} &  - \\
         & & GP-Beta \cite{Song2019} & 3.419 & 0.091 & 2.169 & \textbf{14.892} & \textbf{0.173} & 13.677 & - \\
         & & Var. Scaling \cite{Laves2020,Levi2019} & 3.392 & 0.095 & 2.203 & 15.833 & 0.180 & 13.568 & \textbf{0.131} \\
         & & GP-Normal & 3.392 & 0.095 & 2.205 & 15.820 & 0.180 & 13.568 & \textbf{0.131} \\
         & & GP-Normal (mv.) & 3.434 & 0.097 & 3.356 & 15.342 & 0.200 & 13.353 & 0.133 \\
         & & GP-Cauchy & 3.316 & 0.097 & 1.944 & - & - & 13.262 & - \\ \cline{2-10}
		 & \multirow{7}{*}{COCO} & Uncalibrated & 4.694 & 0.115 & 4.999 & 553.244 & 0.530 & 18.778 & \textbf{0.153} \\
         & & Isotonic Reg. \cite{Kuleshov2018} & \textbf{3.886} & \textbf{0.029} & \textbf{4.534} & 105.343 & 0.113 & \textbf{15.544} & - \\
         & & GP-Beta \cite{Song2019} & 4.169 & 0.142 & 5.093 & \textbf{77.017} & \textbf{0.071} & 16.677 & - \\
         & & Var. Scaling \cite{Laves2020,Levi2019} & 4.204 & 0.167 & 5.606 & 83.653 & 0.072 & 16.815 & 0.207 \\
         & & GP-Normal & 4.204 & 0.167 & 5.606 & 84.367 & 0.072 & 16.815 & 0.207 \\
         & & GP-Normal (mv.) & 4.236 & 0.158 & 6.233 & 82.074 & 0.087 & 16.455 & 0.194 \\ 
         & & GP-Cauchy & 3.936 & 0.043 & 4.716 & - & - & 15.745 & - \\
         \hline
	\end{tabular}
\end{table}
\begin{figure}[t!]
	\centering
\begin{tikzpicture}

\definecolor{color0}{rgb}{0.12156862745098,0.466666666666667,0.705882352941177}
\definecolor{color1}{rgb}{0.0,0.597656,0.296875}
\definecolor{color2}{rgb}{1.0,0.5,0.0}
\definecolor{color3}{rgb}{0.5,0.0,0.25}
\definecolor{color4}{rgb}{0.0,0.5,0.5}
\definecolor{color5}{rgb}{0.0,0.0,0.0}

\begin{axis}[
width=\linewidth,
height=0.4\linewidth,
legend cell align={left},
legend style={
  fill opacity=0.8,
  draw opacity=1,
  text opacity=1,
  at={(0.7,0.75)},
  anchor=north west,
  draw=white!80!black,
  nodes={scale=0.8, transform shape},
},
legend image post style={scale=0.8},
tick align=outside,
tick pos=left,
title={Reliability diagram for $x$ coordinate},
x grid style={white!69.0196078431373!black},
xlabel={Expected quantile $\quantile$},
xmajorgrids,
xmin=0, xmax=1,
xtick style={color=black},
xtick={0,0.2,0.4,0.6,0.8,1},
xticklabels={0.0,0.2,0.4,0.6,0.8,1.0},
y grid style={white!69.0196078431373!black},
ylabel={Observed frequency},
ymajorgrids,
ymin=0, ymax=1,
ytick style={color=black},
ytick={0,0.2,0.4,0.6,0.8,1},
yticklabels={0.0,0.2,0.4,0.6,0.8,1.0}
]
\addplot [semithick, color0, solid, mark=*, mark size=1.5, mark options={solid}]
table {%
0.05 0.130131616949292
0.1 0.23880843338697
0.15 0.331889260221975
0.2 0.414201427569419
0.25 0.486322564673846
0.3 0.551388373148492
0.35 0.60704707678343
0.4 0.658951190328836
0.45 0.703676197549202
0.5 0.746214465486653
0.55 0.784090440236003
0.6 0.815983826381153
0.65 0.843833807814498
0.7 0.868094236085324
0.75 0.890869332013038
0.8 0.911540207121343
0.85 0.929529232165697
0.9 0.947683294137063
0.95 0.964764616082849
};
\addlegendentry{Uncalibrated}

\addplot [semithick, dash pattern=on 1pt off 3pt on 3pt off 3pt, color1, mark=*, mark size=1.5, mark options={solid}]
table {%
	0.05 0.0503775219705409
	0.1 0.125964434542229
	0.15 0.160374633824318
	0.2 0.198333127037175
	0.25 0.228493625448694
	0.3 0.293188100837562
	0.35 0.350909766060156
	0.4 0.405165655815489
	0.45 0.454387919296943
	0.5 0.498452778809259
	0.55 0.554482815529975
	0.6 0.607005817551677
	0.65 0.650988158600487
	0.7 0.700829310558237
	0.75 0.755580311094607
	0.8 0.805751536906383
	0.85 0.853612245739984
	0.9 0.903907249247019
	0.95 0.950158848042249
};
\addlegendentry{Isotonic Regression}

\addplot [semithick, dotted, color2, mark=diamond*, mark size=1.5, mark options={solid}]
table {%
	0.05 0.0833849073730247
	0.1 0.168255147089161
	0.15 0.238107026447168
	0.2 0.312992532079053
	0.25 0.367867310310682
	0.3 0.430952675661179
	0.35 0.484012047695672
	0.4 0.533110533481867
	0.45 0.573750876758675
	0.5 0.621281511738251
	0.55 0.66006518958617
	0.6 0.706853158394191
	0.65 0.738416470685316
	0.7 0.774064446919998
	0.75 0.812971902463176
	0.8 0.841935883153856
	0.85 0.874695713165821
	0.9 0.905268803894872
	0.95 0.937739819284565
};
\addlegendentry{GP-Beta}

\addplot [semithick, color3, opacity=1.0, dashed, mark=square*, mark size=1.5, mark options={solid}]
table {%
	0.05 0.117506292032842
	0.1 0.216487189008541
	0.15 0.302966538763048
	0.2 0.37900730288402
	0.25 0.450715847670916
	0.3 0.513842472253167
	0.35 0.568222139703759
	0.4 0.617815736270991
	0.45 0.664644964310765
	0.5 0.706523084540166
	0.55 0.746296983950159
	0.6 0.782275034038866
	0.65 0.813343235548954
	0.7 0.841399513141065
	0.75 0.865701200643644
	0.8 0.889796591987457
	0.85 0.911870280975368
	0.9 0.932004786070883
	0.95 0.953707141973016
};
\addlegendentry{Variance Scaling}

\addplot [semithick, color4, opacity=1.0, dash pattern=on 3pt off 6pt on 6pt off 6pt, mark=triangle*, mark size=1.5, mark options={solid}]
table {%
	0.05 0.117176218178818
	0.1 0.215785782068738
	0.15 0.302223872591492
	0.2 0.377728266699674
	0.25 0.449230515327805
	0.3 0.51268721376408
	0.35 0.566447992738375
	0.4 0.615876552378595
	0.45 0.66299459504064
	0.5 0.704913974501795
	0.55 0.744811651607047
	0.6 0.780954738622767
	0.65 0.811651607047077
	0.7 0.840574328506003
	0.75 0.864339645995792
	0.8 0.889053925815901
	0.85 0.911375170194331
	0.9 0.931262119899328
	0.95 0.953335808887239
};
\addlegendentry{GP-Variance}

\addplot [semithick, color5, mark=*, mark size=1.5, mark options={solid}]
table {%
	0.05 0.0595370714197302
	0.1 0.106118744068985
	0.15 0.152782935181747
	0.2 0.199447126294508
	0.25 0.247184057432851
	0.3 0.292156620043735
	0.35 0.337129182654619
	0.4 0.383545818376862
	0.45 0.433015637248834
	0.5 0.48318686306061
	0.55 0.53793786359698
	0.6 0.592276271815819
	0.65 0.652514750175352
	0.7 0.715435078598836
	0.75 0.787473697239757
	0.8 0.853364690349466
	0.85 0.915707389528407
	0.9 0.966208689194207
	0.95 0.993728596773528
};
\addlegendentry{GP-Cauchy}

\addplot [semithick, red, dashed]
table {%
0 0
1 1
};
\addlegendentry{Perfect Calibration}
\end{axis}

\end{tikzpicture}
	\caption{Reliability diagram which shows the fraction of ground-truth samples that are covered by the predicted quantile for a certain $\quantile$. The \fasterrcnn{} underestimates the prediction uncertainty. The best calibration performance w.r.t. \textit{quantile calibration} is reached by \textit{Isotonic Regression} \cite{Kuleshov2018} and \textit{GP-Cauchy}.}
	\label{fig:quantiles}
\end{figure}
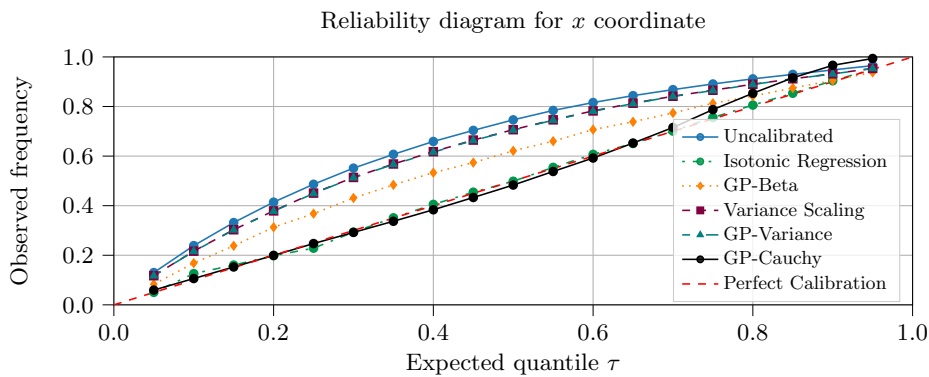 \\

\textbf{Observations.} The experiments show that the simple \textit{Isotonic Regression} \cite{Kuleshov2018} achieves the best results regarding \textit{quantile calibration} (cf. metrics QCE and Pinball loss) as well as for the NLL. In contrast, the calibration methods \textit{Variance Scaling} and \textit{GP-Normal} only lead to slight or no improvements w.r.t. to \textit{quantile calibration} but achieve better results for UCE and ENCE. These metrics measure the miscalibration by asserting normal distributions which leads to the assumption that these methods are beneficial if normal distributions are required for subsequent processes. To get a better insight, we further inspect the error distribution as well as the reliability diagrams for quantile regression in Fig.~\ref{fig:errordist} and \ref{fig:quantiles}, respectively. 
On the one hand, our experiments show that the estimation error of the base networks is centered around 0 but also has heavier tails compared to a representation by a normal distribution. In these cases, the error distribution can be better expressed in terms of a Cauchy distribution (\textit{GP-Cauchy}). On the other hand, we cannot find a strong connection between the error and the bounding box position and/or shape. This can be seen by comparing the calibration performance of \textit{GP-Normal} and the more simple \textit{Variance Scaling} method that does not take additional distribution information into account. Therefore, marginal recalibration in terms of the basic \textit{quantile calibration} is sufficient in this case which explains the success of the simple \textit{Isotonic Regression} calibration method. Although the \textit{GP-Beta} is also not bounded to any parametric output distribution and should also lead to \textit{quantile calibration}, it is restricted to the beta calibration family of functions which limits the representation power of this method. In contrast, \textit{Isotonic Regression} calibration is more flexible in its shape for the output distribution.
However, applications such as Kalman filtering require a normal distribution as input. Although \textit{Isotonic Regression} achieves good results with respect to \textit{quantile calibration}, it only yields poor results in terms of \textit{variance calibration} (cf. UCE and ENCE). In contrast, \textit{Variance Scaling} and especially the multivariate \textit{GP-Normal} achieve the best results since these methods are designed to represent the error distribution as a Gaussian. In a nutshell, if the application requires to extract statistical moments (e.g. mean and variance) for subsequent processing, \textit{Isotonic Regression} is not well suited since the calibration properties are aligned w.r.t. the quantiles and not a parametric representation. In comparison, a direct calibration with parametric models (e.g. \textit{GP-Normal}) leads to better properties regarding the statistical moments.

Therefore, we conclude that the error distribution is represented best by the simple \textit{Isotonic Regression} when working with quantiles, and by \textit{Variance Scaling} and multivariate \textit{GP-Normal} when working with normal distributions. In our experiments, the error distribution rather follows a Cauchy distribution than a normal distribution. Furthermore, we could not find a strong connection between position information and the error distribution. Thus, marginal recalibration is sufficient in this case. Therefore, the choice of the calibration method depends on the use case: if the quantiles are of interest, we advise the use of \textit{Isotonic Regression} or \textit{GP-Beta}. If a representation in terms of a continuous parametric distribution is required, the \textit{GP-Cauchy} or \textit{GP-Normal} achieve reasonable results. If a normal distribution is required, we strongly suggest the use of the multivariate \textit{GP-Normal} recalibration method as it yields the best calibration results for the statistical moments and is well suited for a Gaussian representation of the uncertainty.

	\section{Conclusion}
For environment perception in the context of autonomous driving, calibrated spatial uncertainty information is of major importance to reliably process the detected objects in subsequent applications such as Kalman filtering.
In this work, we give a brief overview over the existing definitions for regression calibration.
Furthermore, we extend the GP recalibration framework by \cite{Song2019} and use parametric probability distributions (Gaussian, Cauchy) as calibration output on the one hand. This is an advantage when subsequent processes require a certain type of distribution, i.e., a Gaussian distribution for Kalman filtering. On the other hand, we use this framework to perform \textit{covariance estimation} of independently learnt distribution for multivariate regression tasks. This enables a post-hoc introduction of (conditional) dependencies between multiple dimensions and further improves the uncertainty representation.

We adapt the most relevant recalibration techniques and compare them to our parametric GP methods to recalibrate the spatial uncertainty of probabilistic object detection models \cite{He2019,Hall2020,Harakeh2020}. Our examinations show that common probabilistic object detection models are too underconfident in their spatial uncertainty estimates. 
Furthermore, the distribution of the prediction error is more similar to a Cauchy distribution than to a normal distribution (cf. Fig.~\ref{fig:errordist}). We cannot find significant correlations between prediction error and position information. Therefore, the definition for \textit{quantile calibration} is sufficient to represent the uncertainty of the examined detection models. This explains the superior performance of the simple \textit{Isotonic Regression} method which performs a mean- and variance-agnostic (marginal) recalibration of the cumulative distribution functions. In contrast, we can show that our multivariate \textit{GP-Normal} method achieves the best results in terms of \textit{variance calibration} which is advantageous when normal distributions are required as output.

\section*{Acknowledgement}
The authors gratefully acknowledge support of this work by Elektronische Fahrwerksysteme GmbH, Gaimersheim, Germany. The research leading to the results presented above are funded by the German Federal Ministry for Economic Affairs and Energy within the project “KI Absicherung – Safe AI for automated driving".

	\clearpage
	%
	%
	\bibliographystyle{splncs04}
	\bibliography{bib/bibliography}
\end{document}